\documentclass[11pt]{article}

\usepackage[preprint]{acl}

\usepackage{times}
\usepackage{latexsym}
\usepackage{mathptmx}
\usepackage{float}
\usepackage{todonotes}
\usepackage{outlines}
\usepackage{algpseudocode}
\usepackage{booktabs}
\usepackage{subcaption}
\usepackage{amsmath}
\usepackage{bm}
\DeclareMathAlphabet{\mathbfit}{OT1}{ptm}{b}{it}
\usepackage{listings}
\usepackage{makecell}
\usepackage{amsfonts}
\usepackage{tikz}
\usetikzlibrary{shapes.geometric}

\usepackage[T1]{fontenc}

\usepackage[utf8]{inputenc}

\usepackage{microtype}

\usepackage{inconsolata}

\usepackage{graphicx}

\usepackage{enumitem}
\setitemize{noitemsep,topsep=0pt,parsep=0pt,partopsep=0pt}

\usepackage{latexml}
\iflatexml
\floatstyle{ruled}
\newfloat{promptfloat}{H}{lop}[section]
\floatname{promptfloat}{Prompt}
\makeatletter
\define@key{promptbox}{label}{\def\promptbox@label{#1}}
\newenvironment{promptbox}[2][]{%
  \let\promptbox@label\relax
  \setkeys{promptbox}{#1}%
  \begin{promptfloat}[H]%
  \caption{#2}%
  \ifx\promptbox@label\relax\else\label{\promptbox@label}\fi}
  {\end{promptfloat}}
\makeatother
\else
\usepackage[most]{tcolorbox}
\newtcolorbox[auto counter, number within=section]{promptbox}[2][]{%
  enhanced, breakable, colback=black!3, colframe=black!60,
  colbacktitle=black!60, fonttitle=\bfseries\small,
  title={Prompt~\thetcbcounter: #2}, #1}
\fi
\iflatexml
\newenvironment{prompt}
  {\par\noindent\ttfamily\small\catcode`\@=9 \catcode`\^^M=\active \promptlines}
  {\par}
\begingroup
\catcode`\^^M=\active %
\gdef\promptlines{\def^^M{\gdef^^M{\newline}}}%
\endgroup
\else
\lstnewenvironment{prompt}[1][]
  {\lstset{basicstyle=\ttfamily\small,
           breaklines=true,
           breakindent=0pt,
           escapechar=@,
           #1}}
  {}
\fi
\newcommand{\role}[1]{\textbf{#1:}}
\newcommand{\tmplarg}[1]{\textcolor{blue}{\{#1\}}}

\title{Activation Steering with Jungian Cognitive Functions}

\author{
  \textbf{Liu Zai\textsuperscript{1}},
  \textbf{Yumeng Wang\textsuperscript{2}\textsuperscript{\dag}},
  \textbf{Junchen Fu\textsuperscript{1}},
  \textbf{Suzan Verberne\textsuperscript{2}},
  \textbf{Joemon M. Jose\textsuperscript{1}}
\\
\\
  \textsuperscript{1}University of Glasgow,
  \textsuperscript{2}Leiden University
}

\begin{document}
\setlength{\abovedisplayskip}{3pt}
\setlength{\belowdisplayskip}{3pt}
\setlength{\abovedisplayshortskip}{0pt}
\setlength{\belowdisplayshortskip}{3pt}
\maketitle
\makeatletter
\iflatexml
\else
\ifacl@anonymize
\else
\begingroup
\renewcommand{\thefootnote}{\dag}
\footnotetext{Corresponding author.}
\endgroup
\fi
\fi
\makeatother
\begin{abstract}
Controlling combinations of personality-related tendencies is important for adapting large language models to diverse interaction needs. But successful single-trait steering does not establish how to construct interventions for joint targets, or whether existing trait directions adequately represent them. 
We introduce a target-first framework that directly extracts steering directions from joint personality targets and then examines their decomposition into individual directions. 
To support this framework, we construct NarrationDB, linking character self-narrations to automatically evaluated joint scores across eight Jungian-inspired functions, a correspondence not directly available in existing Big Five- or MBTI-oriented resources. 
Experiments on three models demonstrate questionnaire-score controllability across the eight functions and coordinated changes in a tested two-function target. Directly extracted joint directions are better approximated by their corresponding individual directions than by unrelated pairings, but retain substantial residual components. 
Our work provides a method and resource for investigating joint personality control, with evidence that individual steering directions capture part, but not all, of the structure of joint targets.
\end{abstract}

\section{Introduction}

\begin{figure*}[t]
\centering
\includegraphics[width=0.9\linewidth]{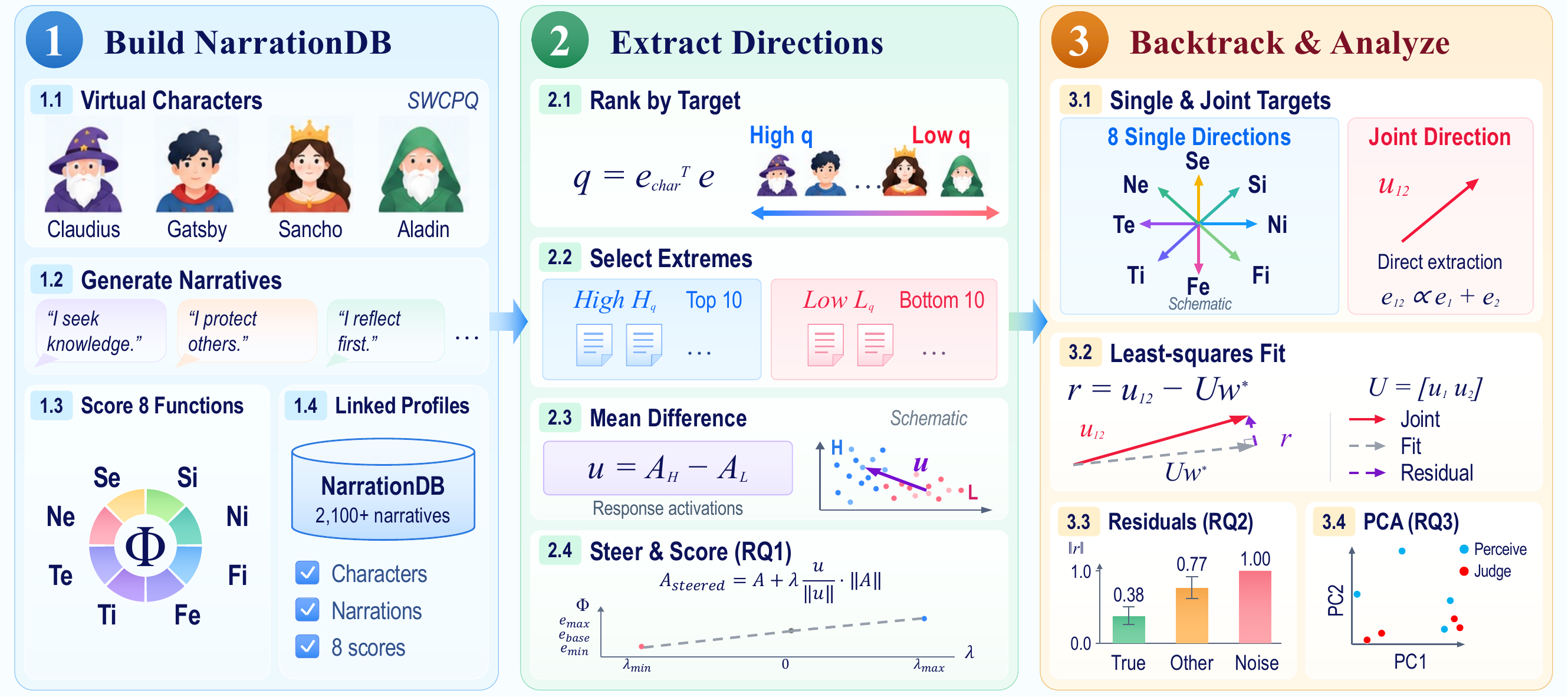}
\caption{Overview of our framework. We build \texttt{NarrationDB} by scoring role-played character narrations on eight Jungian functions, extract single- and joint-target steering directions from its high- and low-scoring characters, and use \texttt{backtracking} to test how well single-function directions explain joint ones.}
\label{fig:pipeline}
\end{figure*}

Adapting large language models (LLMs) to diverse interaction needs often requires multiple psychological functions to work together. Each function in the Jungian eight-function model \cite{jung1971psychological} is Thinking (\(\mathit{T}\)), Feeling (\(\mathit{F}\)), Sensing (\(\mathit{S}\)), or Intuition (\(\mathit{N}\)), in an Introverted (\(\mathit{i}\)) or Extroverted (\(\mathit{e}\)) form. In Beebe's model \cite{beebe}, a dominant and an auxiliary function jointly characterize a thinking style. For example, a brainstorming assistant may generate creative possibilities (\(\mathit{Ne}\)) and critically examine their underlying logic (\(\mathit{Ti}\)), developing ideas that are both innovative and coherent. Activation steering offers a way to influence such functions without updating model parameters by intervening on intermediate activations \cite{frising2026,panickssery2024,chen2025}. However, constructing an intervention for a joint target requires more than demonstrating that each constituent trait can be steered separately \cite{bhandari2026interfere}. It requires understanding whether existing single-trait directions capture the joint target or leave relevant components unexplored.

Research on personality steering has established trait-aligned directions using continuous Big Five scores \cite{frising2026}, while also showing that geometric independence does not guarantee independent behavioral effects \cite{bhandari2026interfere}. These findings motivate a construction problem: \textbf{How can we derive effective steering directions for joint personality targets?} A directly extracted joint direction provides an empirical reference for examining this problem. It allows the explanatory reach of individual directions to be measured, including components outside their span.

We address this problem through a \textit{target-first framework} (Figure~\ref{fig:pipeline}): each target, single or joint, is specified as a score before its direction is extracted. We use these eight functions as separately scored behavioral dimensions. The functions describe tendencies in attending to information and making judgments, providing interpretable targets such as exploring possibilities and emphasizing logical consistency. Our evaluation assigns each function a questionnaire score without forcing functions into mutually exclusive pairs. Studying their combinations requires text associated with all eight scores. Resources such as BIG5-CHAT \cite{li2025big5} provide high/low Big Five trait labels, while continuous Big Five profiles and MBTI types do not directly supply this eight-function correspondence. We therefore construct \texttt{NarrationDB}, linking more than \( 2{,}100 \) character self-narrations to multidimensional profiles. Human-rated character attributes inform narration generation, and a role-conditioned LLM supplies the function scores.

\texttt{NarrationDB} enables the same extraction procedure to be applied to different targets. For a single function, we select high- and low-scoring characters and extract a difference-in-means direction from their response activations. For a joint target, we select characters using a combined score and extract a new direction directly. We then perform \texttt{backtracking}, testing how well this joint direction can be reconstructed from its constituent single-function directions. The joint direction is unconstrained by their span during extraction, making its linear decomposability an empirical question.

Experiments on Llama-3.1-8B-Instruct \cite{llama3}, Qwen3-8B \cite{qwen3}, and Gemma-3-4B-IT \cite{gemma3} show questionnaire-score controllability across all eight functions and coordinated changes in a tested two-function target. Across twelve joint targets, corresponding single-function subspaces approximate directly extracted directions better than unrelated pairings, although substantial residuals remain. This supports approximate linear compositional structure while identifying components missed by single-function representations. Our claims concern measured scores and extracted directions; whether linear approximations reproduce the joint intervention's behavioral effects remains a separate question.

We address the following research questions:

\begin{outline}
  \1 \textbf{RQ1}: Do cognitive-process personality dimensions correspond to steerable directions in LLM activation space?
  \1 \textbf{RQ2}: To what extent can joint-target steering directions be explained by combinations of single-function directions?
  \1 \textbf{RQ3}: What geometric patterns emerge among the eight cognitive-function steering directions, and how do they relate to psychological theory?
\end{outline}

Our contributions are:

\begin{outline}
  \1 \textbf{A target-first framework for joint personality steering.} We directly extract joint-target directions and use backtracking to assess how well individual directions explain them.
  \1 \textbf{NarrationDB, a resource for constructing joint targets.} Character self-narrations and eight-function profiles support consistent sample selection for single and joint targets.
  \1 \textbf{Evidence of partial linear compositionality.} Single-function and joint-target experiments characterize controllability and reveal both shared linear structure and residual components in joint directions.
\end{outline}

\section{Related Work}

\subsection{Personality in LLMs}

Recent work has applied psychometric inventories to LLMs. \citet{serapio2023personality} found that personality measurements of some models are reliable and valid under specific prompting configurations. \citet{jiang2023mpi} introduced a Big Five-based Machine Personality Inventory and induced target personalities through prompting, and \citet{jiang2024personallm} showed that assigned Big Five personas yield consistent questionnaire scores and trait-related writing. MBTI-style tests have also been applied to LLMs as a rough personality indicator \cite{pan2023mbti}. PsychoBench \cite{huang2024} further extends that evaluation to broader psychological scales. InCharacter \cite{wang2024incharacter} shows that role-playing agents reproduce measurable personalities of their target characters, which motivates our use of character narrations in \texttt{NarrationDB}. Personality can also be shaped through training on trait-labeled data \cite{li2025big5}. These works predominantly adopt the Big Five personality model or dichotomous MBTI types, whereas we score eight Jungian Cognitive Functions separately.

\subsection{Activation Steering}

Activation steering controls behavior by adding directions to intermediate activations \cite{turner2024,panickssery2024,zou2023repe,li2023iti}, building on the hypothesis that high-level concepts are linearly represented \cite{park2024linear,marks2024}. \citet{allbert2025} were among the first to apply activation engineering to personality traits, and \citet{frising2026} situated personality steering within the Big Five model, finding linear directions to be effective probes but weaker steering vectors. Persona Vectors \cite{chen2025} extend such directions to behavioral traits such as sycophancy.

For multiple attributes, existing methods typically combine single-attribute vectors, via orthogonalized vector algebra for personality \cite{feng2026persona} or sparse, mutually orthogonal attribute vectors \cite{nguyen2025matsteer}, and additive composition has been reported to succeed when vectors are approximately orthogonal \cite{kang2026compositional}. In contrast, \citet{bhandari2026interfere} found Big Five steering directions to be geometrically dependent. Rather than assuming that single-trait directions form an adequate basis, we extract joint-target directions directly and use \texttt{backtracking} to measure how much of them the single-function directions explain.

\section{Methodology}

\subsection{Problem Formulation}

Let the semantic space be defined as
\[
  E = \mathbb{R}^{8}
\]
where each dimension corresponds to one of the eight Jungian Cognitive Function scores. Let \(p\) denote an input prompt, and let \(\Phi\) denote the semantic evaluation procedure that maps a model response to its corresponding point in \(E\).

Let \(M(p;u)\) denote the response generated by the LLM for prompt \(p\) under steering intervention \(u\), where \(u\) belongs to an admissible steering-intervention space \(\mathcal{U}\). The unsteered semantic position is defined as
\[
  x_0 = \Phi(M(p;0))
\]
while the semantic position obtained under steering intervention \(u\) is
\[
  x(u) = \Phi(M(p;u))
\]
The resulting semantic displacement induced by steering is therefore
\[
  \Delta(u;p) = x(u)-x_0
\]
For a target semantic direction \(e \in E\), normalized to unit length, the directional semantic effect of a steering intervention is defined as
\[
  J_e(u;p) = \Delta(u;p)^\top e
\]
This quantity measures the component of the semantic displacement in the target direction \(e\). When \(e\) corresponds to one of the standard basis vectors of \(E\), \(J_e(u;p)\) is equivalent to the change in the corresponding Jungian Cognitive Function score.

Under this formulation, the ideal steering intervention for a single semantic direction is given by
\[
u_e^* = \arg\max_{u\in\mathcal{U}} \Delta(u;p)^\top e
\]
The expression represents an idealized optimization objective. In the empirical setting, the optimizer is approximated by searching over a finite set of admissible steering interventions and evaluating their resulting semantic effects.

\subsection{Target-Conditioned Direction Extraction and Steering}

Steering LLMs toward different Jungian Cognitive Function combinations requires finding prompts that effectively influence positioning in semantic space \( E \), and constructing a dataset of such prompts from which differential clusters related to both isolated and combined semantics can be meaningfully retrieved. \texttt{NarrationDB} in Section~\ref{s:ndb} is constructed to meet this purpose.

We then use the difference-in-means direction \cite{marks2024,arditi2024refusal} to extract the intervention condition \( u_e^* \in U \), which, from another perspective, is the vector in activation space representing the control target \( e \) as defined above. First, we construct a pivot score for each character narration:
\[
  q = e_\mathit{character}^T e
\]
In single-dimensional control experiments, where \( e \) is a one-hot encoding of the target Jungian Function, trivially \( q = e_\mathit{Cf} \). With joint steering targets, \( q \) is the sum of all relevant scores if all steering targets have the same weights and signs, i.e., \( q_{12} = e_{1} + e_{2} \).

To extract the intervention condition \( u_e^* \), we first identify a high-scoring cluster of virtual characters, \( H_q \), and a low-scoring cluster, \( L_q \), in the sample space within \texttt{NarrationDB}. We select the clusters here by sorting all characters by \( q \) and taking the top and bottom \( k \) virtual characters to form the clusters.

We then prompt each character in the cluster with five open-ended questions about personality. We extract the average representation of the response tokens during the decoding phase \cite{chen2025,frising2026} as \( A_{\mathit{character}, t} \), where \( t \) is the index of the open-ended question. The average activation for the character is then extracted as the average activation of all open-ended questions:
\[
  A_\mathit{character} = \frac{1}{T} \sum_{t=1}^{T} A_{\mathit{character}, t}
\]
where \( T \) is the total number of open-ended questions. Similarly, the average latent for the two clusters is extracted as:
\[
  A_{H_q} = \frac{1}{k} \sum_{i=1}^{k} A_{H_q, i}
\]
\[
  A_{L_q} = \frac{1}{k} \sum_{i=1}^{k} A_{L_q, i}
\]
Finally, the activation direction \( u_e^* \) is extracted as the difference between the two clusters' average latent:
\[
  u_e^* = A_{H_q} - A_{L_q}
\]
\citet{belrose2024} shows that \( u_e^* \) is worst-case optimal. Following prior activation-steering work \cite{panickssery2024}, we separate steering direction from steering magnitude by normalizing it to unit length before injection. The extracted steering vector is added to the activation during inference, with a control strength coefficient \( \lambda \):
\[
  A_\mathit{steered} = A + \lambda \frac{u}{\lVert u \rVert} \cdot \lVert A \rVert
\]
\subsection{Joint-Target Steering Analysis via Backtracking}

Previous studies focus primarily on combining existing single-dimensional steering vectors. This approach has a few flaws:
\textbf{(1) Assumption of the existence of a combined semantic direction.}
Semantic directions are assumed to be isolated and stackable.
\textbf{(2) Limited search space.}
The activation space is high-dimensional, and combinations of single vectors typically span only a fraction of the total volume. The target direction might lie outside scanned space.
\textbf{(3) Unobserved relationships remain unidentified.}
Existing approaches can only determine whether a combining method is effective, without investigating whether unobserved relationships exist.

To address these challenges, we extract multi-dimensional activation-space directions from the semantic sample space and examine the potential existence of a linear mapping between these spaces. This method relaxes these assumptions by starting with the target direction in semantic space rather than activation space, forming a target-first framework instead of constructing multi-dimensional activation-space directions from single-dimensional ones. We will refer to this method as \texttt{backtracking}.

For multi-dimensional steering, consider two target directions \(e_1,e_2\in E\). An equal-weight combined semantic direction can be defined as
\[
e_{12} = \frac{e_1+e_2}{\lVert e_1+e_2\rVert}
\]
provided that \(e_1+e_2\neq 0\). The corresponding scalarized multi-dimensional steering objective is then
\[
J_{12}(u;p) = \Delta(u;p)^\top e_{12}
\]
and the directly extracted multi-dimensional steering intervention is
\[
u_{12}^* = \arg\max_{u\in\mathcal{U}} J_{12}(u;p)
\]
The central question of multi-dimensional steering is whether the directly extracted intervention \(u_{12}^*\) can be explained as a combination of the corresponding single-dimensional interventions. Let
\[
u_1^* = \arg\max_{u\in\mathcal{U}} \Delta(u;p)^\top e_1
\]
and
\[
u_2^* = \arg\max_{u\in\mathcal{U}} \Delta(u;p)^\top e_2
\]
denote the independently extracted steering interventions for the two individual semantic directions. A combining function can then be formulated as a mapping
\[
C:\mathcal{U}\times\mathcal{U}\rightarrow\mathcal{U}
\]
with the hypothesis that
\[
\widehat{u}_{12}
=
C(u_1^*,u_2^*)
\]
approximates the directly extracted multi-dimensional intervention \(u_{12}^*\).

The traditional forward-tracking approach \cite{feng2026persona,nguyen2025matsteer,kang2026compositional} begins by hypothesizing a particular form of \(C\), constructs \(\widehat{u}_{12}=C(u_1^*,u_2^*)\), and subsequently evaluates whether the resulting intervention produces the intended multi-dimensional semantic effect. In contrast, the \texttt{backtracking} approach proposed reverses this order. It first identifies \(u_{12}^*\) directly from the semantic objective, independently obtains \(u_1^*\) and \(u_2^*\), and then examines whether a mapping \(C\) can be inferred that relates the observed tuple \( (u_1^*,u_2^*,u_{12}^*) \).

\section{NarrationDB}
\label{s:ndb}

\subsection{Tag-Augmented Narration Generation}

To generate \texttt{NarrationDB}, a Jungian personality dataset for virtual characters, we first instruct Llama-3.1-70B-Instruct \cite{llama3} to role-play as virtual characters in the SWCPQ dataset \cite{jorgenson2020swcpq}, then collect their self-narrations with the prompt listed in Prompt~\ref{p:gen_ndb}.

The generation process is augmented by including personality tags extracted from the SWCPQ dataset \cite{jorgenson2020swcpq}. The SWCPQ dataset contains human-rated scores for pairs of description tags, ranging from \( 0 \) to \( 100 \). For each tag associated with a character, we calculate the margin of error for the score at a confidence level of \( 95\% \), and only the \( 20 \) sliders with the smallest \( \mathit{MOE} \) are considered. Finally, \( 10 \) tags with scores closer to \( 0 \) (for tags on the left-hand side), or \( 100 \) (for tags on the right-hand side), are extracted as the tags for that character. This step selects \( 10 \) tags from \( 2 \times 20 \) tags of most agreeable scores on both sides of the slider. Some examples of the results of this process are listed in Table~\ref{tab:ndb_tags}.

\subsection{Jungian Evaluation}

We extract questionnaire data from an open-source multi-dimensional Jungian evaluation project \cite{felmonon2026} for human subjects and assemble it into questionnaires to assess the Jungian Cognitive Functions of LLMs. We randomly shuffle the questions and divide them into groups. Each group is then passed to target LLMs with a PsychoBench-like \cite{huang2024} prompt prefix (see Prompt~\ref{p:eval}), and the score for each Jungian Cognitive Function (CF) is calculated in the same manner as the original project \cite{felmonon2026}:
\[
  w_\mathit{Cf} = 0.7 \overline{P_\mathit{Cf}} + 0.3 (6 - \overline{N_\mathit{Cf}})
\]
\[
  e_\mathit{Cf} = 100 ( \frac{w_\mathit{Cf} - 1}{4} )
\]
where \( P_\mathit{Cf} \) denotes the set of questions that contribute positively to the current function, and \( N_\mathit{Cf} \) denotes the set of questions that contribute negatively to the current function. \( \overline{P_\mathit{Cf}} \) and \( \overline{N_\mathit{Cf}} \) denote the average scores of questions in set \( P_\mathit{Cf} \) and \( N_\mathit{Cf} \), respectively.

We develop a quantitative analysis protocol to verify that \texttt{NarrationDB} samples are abundant enough to serve as the foundation dataset. We ask Llama-3.1-8B-Instruct \cite{llama3}, Qwen3-8B \cite{qwen3}, and Gemma-3-4B-IT \cite{gemma3} to role-play as the character in a manner consistent with the generated narration in the system prompt as listed in Prompt~\ref{p:eval_rp}. The aforementioned Jungian Cognitive Function test is then performed on the character and the character identity, franchise, narration, and their Jungian Cognitive Function scores are stored in \texttt{NarrationDB}.

\begin{figure}[htb]
\centering
\includegraphics[width=\linewidth]{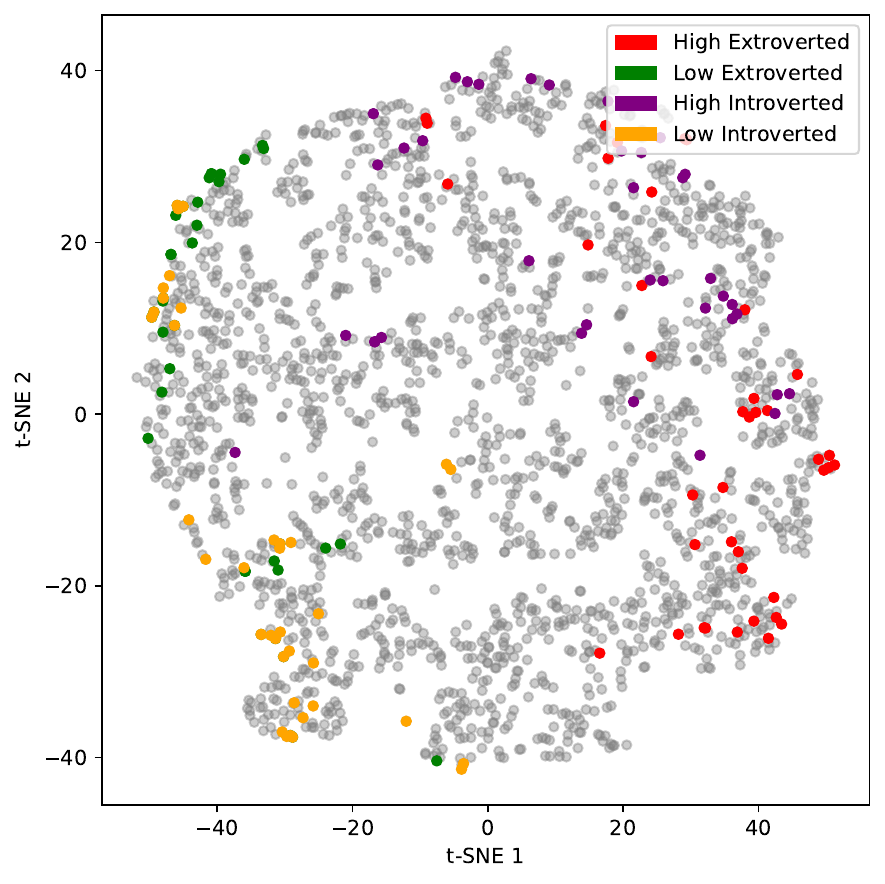}
\caption{t-SNE plot of \texttt{NarrationDB}, with the top 10 items of eight dimensions colored. Colored items are from the same clusters used for activation extraction.}
\label{fig:ndb_tsne}
\end{figure}

The resulting dataset \texttt{NarrationDB} contains \( 2{,}100+ \) virtual characters with diverse personalities, and these characters successfully fill the eight-dimensional space with their rich personalities, as shown in Figure~\ref{fig:ndb_tsne}. This implies meaningful directions in semantic space \( E \) which lays a foundation for activation steering.

\section{Experiments}

\subsection{Experiment Setup}

\begin{table*}[t]
\small
\centering
{
\setlength{\tabcolsep}{2pt}
\footnotesize
\begin{tabular}{l|rrrcccc|rrrcccc|rrrcccc}
\toprule
& \multicolumn{7}{c|}{\textbf{Llama-3.1-8B-Instruct}}
& \multicolumn{7}{c|}{\textbf{Qwen3-8B}}
& \multicolumn{7}{c}{\textbf{Gemma-3-4B-IT}} \\
\midrule
\( \mathbfit{Cf} \)
  & \( \bm{n^{*}} \) & \( \bm{\lambda_\mathbfit{min}} \) & \( \bm{\lambda_\mathbfit{max}} \) & \( \bm{e_\mathbfit{base}} \) & \( \bm{e_\mathbfit{min}} \) & \( \bm{e_\mathbfit{max}} \) & \( \bm{\Delta_\mathbfit{eff}} \)
  & \( \bm{n^{*}} \) & \( \bm{\lambda_\mathbfit{min}} \) & \( \bm{\lambda_\mathbfit{max}} \) & \( \bm{e_\mathbfit{base}} \) & \( \bm{e_\mathbfit{min}} \) & \( \bm{e_\mathbfit{max}} \) & \( \bm{\Delta_\mathbfit{eff}} \)
  & \( \bm{n^{*}} \) & \( \bm{\lambda_\mathbfit{min}} \) & \( \bm{\lambda_\mathbfit{max}} \) & \( \bm{e_\mathbfit{base}} \) & \( \bm{e_\mathbfit{min}} \) & \( \bm{e_\mathbfit{max}} \) & \( \bm{\Delta_\mathbfit{eff}} \) \\
\midrule
Fe
 & 7  & -0.20 &  0.20 & 81.6 & 39.7 & 92.4 & \textbf{52.6}
 & 12 & -0.20 &  0.05 & 77.3 & 60.6 & 80.6 & \underline{19.9}
 & 12 & -0.02 &  0.03 & 71.6 & 57.9 & 81.8 & 23.9 \\
Fi
 & 12 & -0.30 &  0.05 & 89.5 & 69.1 & 91.2 & 22.0
 & 7  & -0.20 &  0.01 & 73.2 & 65.2 & 74.9 &  9.6
 & 11 & -0.05 &  0.03 & 78.9 & 52.2 & 80.0 & 27.8 \\
Ne
 & 12 & -0.20 &  0.10 & 85.6 & 47.7 & 95.9 & 48.2
 & 15 & -0.20 &  0.05 & 69.6 & 59.6 & 75.1 & 15.5
 & 13 & -0.04 &  0.01 & 81.4 & 36.0 & 86.0 & \textbf{49.9} \\
Ni
 & 18 & -0.50 & -0.02 & 88.7 & 70.6 & 89.6 & 19.0
 & 14 & -0.20 &  0.20 & 69.2 & 62.8 & 75.3 & 12.5
 & 14 & -0.05 &  0.05 & 64.1 & 56.4 & 70.9 & 14.4 \\
Se
 & 12 & -0.20 &  0.10 & 80.5 & 51.7 & 89.0 & 37.2
 & 10 & -0.05 &  0.30 & 71.5 & 66.9 & 80.9 & 13.9
 & 16 & -0.05 &  0.03 & 72.5 & 41.9 & 80.3 & \underline{38.4} \\
Si
 & 7  & -0.50 &  0.20 & 74.7 & 33.7 & 85.4 & \underline{51.7}
 & 11 & -0.30 &  0.20 & 76.0 & 68.2 & 84.3 & 16.0
 & 10 & -0.03 &  0.02 & 76.0 & 65.4 & 83.4 & 18.0 \\
Te
 & 8  & -0.30 &  0.20 & 78.2 & 69.8 & 81.5 & 11.6
 & 14 & -0.30 &  0.30 & 77.1 & 62.0 & 87.1 & \textbf{25.0}
 & 12 & -0.03 &  0.03 & 71.6 & 64.3 & 79.3 & 15.0 \\
Ti
 & 5  & -0.30 &  0.02 & 87.9 & 73.2 & 91.8 & 18.5
 & 6  & -0.30 &  0.20 & 81.7 & 69.3 & 85.8 & 16.4
 & 13 & -0.03 &  0.05 & 78.2 & 71.1 & 82.0 & 10.9 \\
\bottomrule
\end{tabular}
}

\caption{Verification of steering effectiveness for the eight cognitive function categories defined by Jung's typology (Appendix~\ref{a:jung}). \( n^* \) is the most effective layer. \( e_\mathit{base}, e_\mathit{min}, e_\mathit{max} \) are base/smallest/largest score for the steered function. \( \Delta_\mathit{eff} \) is the difference of the smallest and largest score obtained. }
\label{tab:steer_params}
\end{table*}

\paragraph{Extraction of activation directions.}
To extract single-dimensional activation directions for one Jungian Cognitive Function, we first sort \texttt{NarrationDB} by \( e_\mathit{Cf} \), then take the top and bottom \( k=10 \) virtual characters to form the high- and low-score clusters. For extraction of joint semantic directions, \texttt{NarrationDB} is instead sorted with the sum of two scores \( q_{12}=e_1 + e_2 \). Each character in the cluster is then prompted with \( T=5 \) open-ended questions related to personalities, as listed in Appendix~\ref{a:oe}. We extract the average activation of the response tokens during the decoding phase \cite{chen2025,frising2026} after each layer.

\paragraph{Parameter sweep and effectiveness evaluation.}
We first perform a coarse layer scan to identify the most effective layer: we sweep the steering strength \( \lambda \) from \( -1 \) to \( +1 \) after layer \( n \) and record \( \Delta_\mathit{Cf} \), then find the most effective layer \( n^* \) which maximizes \( \Delta_\mathit{eff} = \Delta_\mathit{max} - \Delta_\mathit{min} \). We then perform a fine-grained sweep of the steering strength \( \lambda \) at the most effective layer \( n^* \) to further max out \( \Delta_\mathit{eff} \). We monitor instruction-following capabilities throughout the process; if the task completion rate drops below \( 80\% \) for a steering condition, the result will be discarded. Full details of the sweep are given in Appendix~\ref{a:sweep}.

\paragraph{Assess the linear residual of joint directions via \texttt{backtracking}.}
At this point, we have retrieved and verified the effectiveness of \( (u_1,u_2,u_{12}) \), where \( u_1 \) and \( u_2 \) are single-dimensional steering vectors and \( u_{12} \) is the steering vector for the combined semantic direction. To test if the hypothesized linear relationship exists:
\[
  C: u_{12} = w_1 u_1 + w_2 u_2
\]
We solve the linear system using the Least-Squares method:
\[
  w^* = \arg\min_{w \in \mathbb{R}^2} \, \lVert u_{12} - U w \rVert_2^2
\]
and measure the residual:
\[
  r = u_{12} - w_1^* u_1 - w_2^* u_2 
\]
We then compare the residual with:
\begin{outline}
  \1 \textbf{Unrelated semantics.} Swapping \( u_{12} \) with \( u_{34} \) which is unrelated to \( u_1 \) and \( u_2 \).
  \1 \textbf{White noise.} Swapping \( u_{12} \) with a random white noise vector.
\end{outline}

\subsection{Steering Effectiveness (RQ1)}

\begin{table*}[t]
\small
\centering
\begin{tabular}{l|rrl|rrl|rrl}
\toprule
  & \multicolumn{3}{c|}{\textbf{Llama-3.1-8B-Instruct}}
  & \multicolumn{3}{c|}{\textbf{Qwen3-8B}}
  & \multicolumn{3}{c}{\textbf{Gemma-3-4B-IT}} \\
\midrule
  \textbf{Combination}
  & \( \bm{w_1} \) & \( \bm{w_2} \) & \( \bm{\lVert r \rVert_2} \)
  & \( \bm{w_1} \) & \( \bm{w_2} \) & \( \bm{\lVert r \rVert_2} \)
  & \( \bm{w_1} \) & \( \bm{w_2} \) & \( \bm{\lVert r \rVert_2} \) \\
\midrule
Ti + Fe
 & 0.76 & 0.53 & 0.38
 & 0.79 & 0.62 & 0.36
 & 1.89 & 1.00 & 0.19 \\
Ti + Ne
 & 0.93 & 0.25 & 0.37
 & 1.04 & 0.48 & 0.40
 & 1.09 & 0.11 & 0.06 \\
Ti + Se
 & 0.78 & 0.52 & 0.36
 & 0.86 & 0.53 & 0.36
 & 1.05 & 0.11 & 0.08 \\
Fi + Te
 & 0.21 & 0.80 & 0.49
 & 0.17 & 0.90 & 0.33
 & 0.02 & 1.02 & 0.08 \\
Fi + Ne
 & 0.48 & 0.58 & 0.32
 & 0.41 & 0.67 & 0.31
 & 0.46 & 0.56 & 0.20 \\
Fi + Se
 & 0.65 & 0.45 & 0.53
 & 0.50 & 0.74 & 0.42
 & 0.22 & 0.77 & 0.44 \\
Ni + Te
 & 0.63 & 0.43 & 0.24
 & 0.41 & 0.66 & 0.19
 & 0.31 & 0.70 & 0.05 \\
Ni + Fe
 & 0.82 & 0.39 & 0.40
 & 0.81 & 0.48 & 0.41
 & 1.48 & 0.54 & 0.19 \\
Ni + Se
 & 0.73 & 0.44 & 0.29
 & 0.70 & 0.56 & 0.30
 & 1.05 & 0.15 & 0.08 \\
Si + Te
 & 0.34 & 0.69 & 0.25
 & 0.20 & 0.81 & 0.27
 & 0.07 & 0.97 & 0.07 \\
Si + Fe
 & 0.63 & 0.56 & 0.29
 & 0.66 & 0.56 & 0.29
 & 0.19 & 0.99 & 0.19 \\
Si + Ne
 & 0.75 & 0.37 & 0.65
 & 0.86 & 0.40 & 0.60
 & 0.10 & -0.69 & 0.68 \\
\midrule
Semantic Addition
 & -- & -- & 0.38 \( \pm 0.12 \)
 & -- & -- & 0.35 \( \pm 0.10 \)
 & -- & -- & 0.19 \( \pm 0.19 \) \\
Unrelated Semantics
 & -- & -- & 0.77 \( \pm 0.15 \)
 & -- & -- & 0.77 \( \pm 0.15 \)
 & -- & -- & 0.38 \( \pm 0.20 \) \\
White Noise
 & -- & -- & 1.00
 & -- & -- & 1.00
 & -- & -- & 1.00 \\
\bottomrule
\end{tabular}

\caption{Results for solving multi-dimensional linear systems using the Least-Squares method and measuring the residual from the effective steering vector. \( w_1 \) and \( w_2 \) are the fitted weights of the first and second function directions in each combination, and \( \lVert r \rVert_2 \) is the L2 norm of the residual left in the unit-length joint direction.}
\label{tab:lin_residuals}
\end{table*}

\begin{figure}[t]
\centering
\includegraphics[width=\linewidth]{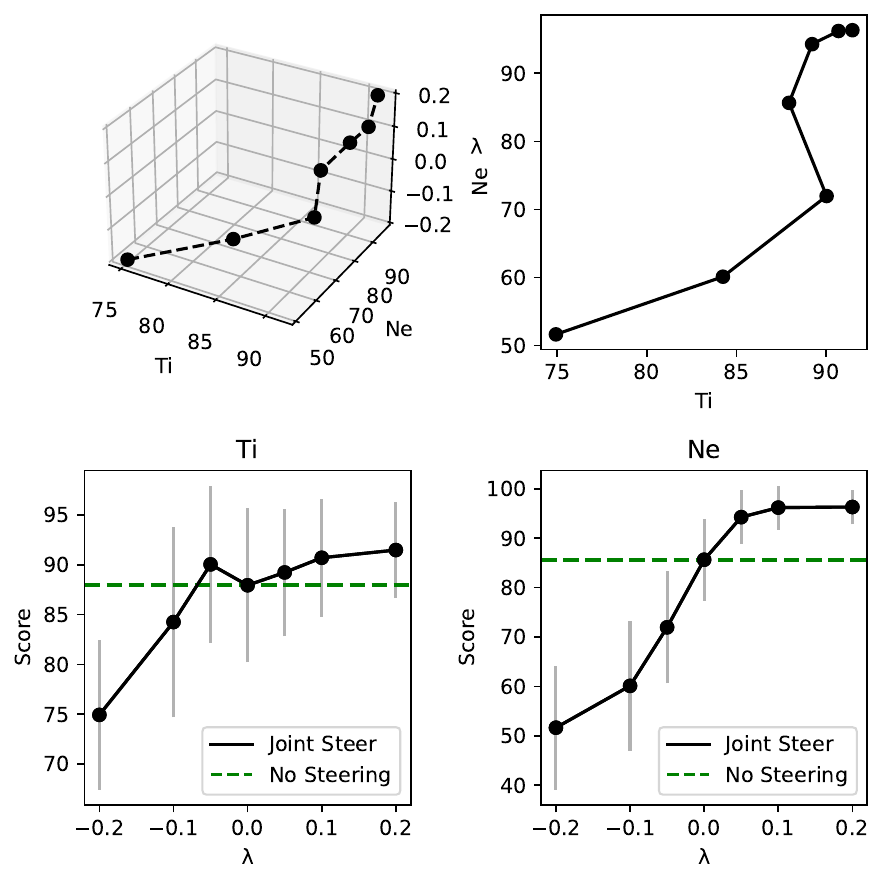}
\caption{Joint-target steering on Llama-3.1-8B-Instruct with a single direction extracted for \( \mathit{Ti}+\mathit{Ne} \). Top panels show trajectory of both scores over \( \lambda \) and its \( \mathit{Ti} \)--\( \mathit{Ne} \) projection. Bottom panels show each score against \( \lambda \), with the unsteered score as reference. Error bars are standard deviations.}
\label{fig:md_ti_ne}
\end{figure}

We report the single-function results in Table~\ref{tab:steer_params}, where steering shifts the targeted score for all eight functions on all three models. Llama-3.1-8B-Instruct is the most steerable overall, with \( \Delta_\mathit{eff} \) around \( 50 \) for \( \mathit{Fe} \), \( \mathit{Si} \) and \( \mathit{Ne} \). Gemma-3-4B-IT reaches above \( 38 \) on \( \mathit{Ne} \) and \( \mathit{Se} \) but remains below \( 28 \) elsewhere. Qwen3-8B is the least responsive, with all functions at or below \( 25 \). We therefore conduct the joint-target case study on Llama-3.1-8B-Instruct (Figure~\ref{fig:md_ti_ne}), where a single vector extracted for \( \mathit{Ti}+\mathit{Ne} \) moves both scores.

These ranges indicate strong controllability. On Llama-3.1-8B-Instruct and Gemma-3-4B-IT, the best functions span roughly half of the score scale, and even the weakest function on each model shifts by about \( 10 \) points. However, the steering effect is asymmetric: the drop from \( e_\mathit{base} \) to \( e_\mathit{min} \) accounts for most of the summed \( \Delta_\mathit{eff} \) per model, while no function rises more than \( 10.8 \) points above its baseline. Failures are concentrated at this upper end. For example, on Llama, \( \mathit{Fi} \) and \( \mathit{Ni} \) gain less than \( 2 \) points, and the highest \( \mathit{Ni} \) score is reached at a negative \( \lambda \).

Our answer to RQ1 is yes. Cognitive-process personality dimensions do correspond to steerable directions in LLM activation space. For all eight Jungian Cognitive Functions and all three models, a difference-in-means steering vector extracted from \texttt{NarrationDB} provides overall effective control of the targeted function, and the same procedure applied to a combined score yields a direction that changes \( \mathit{Ti} \) and \( \mathit{Ne} \) together. Control is strongest when lowering scores; raising them is limited by the high baselines of the unsteered models. The attainable range depends more on the model than on the function.

\subsection{Linear Residuals of Joint-Target Activation Directions (RQ2)}

\begin{figure}[t]
\centering
\includegraphics[width=\linewidth]{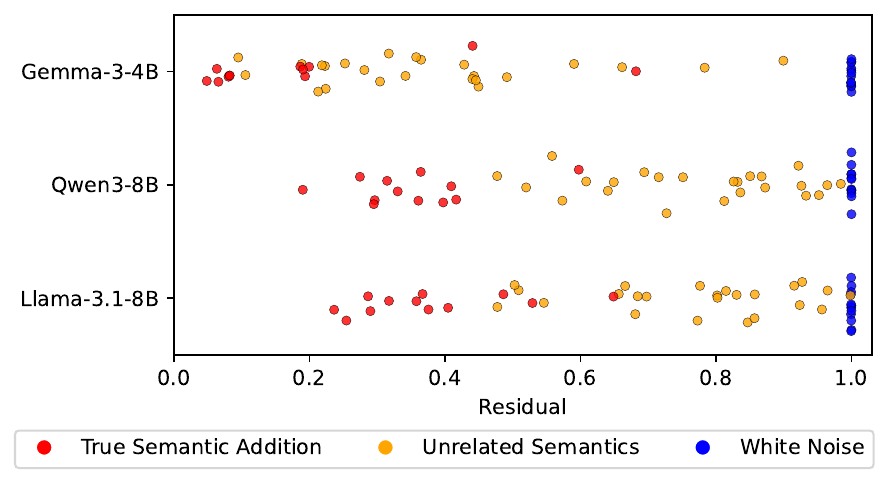}
\caption{Residuals of joint-target directions after least-squares projection onto the plane of their two single-function directions, compared with the same procedure applied to unrelated semantic directions and white-noise vectors.}
\label{fig:residual_scatter}
\end{figure}

We apply \texttt{backtracking} to twelve joint targets, each pairing an introverted with an extroverted function (Table~\ref{tab:lin_residuals}, Figure~\ref{fig:residual_scatter}). On every model, true semantic addition leaves the smallest residuals, and unrelated semantics remain below the white-noise baseline. Residuals near zero occur only on Gemma-3-4B-IT for six targets. Across models, \( \mathit{Ni}+\mathit{Te} \) is consistently the best approximated and \( \mathit{Si}+\mathit{Ne} \) consistently the worst, the latter also being the only target with a negative fitted weight on Gemma.

If a joint direction were an exact linear combination of its constituents, its residual would vanish. White noise leaves a residual of \( 1 \), as a random vector in a high-dimensional space has almost no component in any fixed plane. Unrelated semantics provide a stricter reference: their residuals fall well below \( 1 \), indicating that directions extracted for different functions share a common component, as expected if features are represented non-orthogonally \cite{elhage2022superposition}. Against this reference, true semantic addition halves the residual on every model, so the constituent directions explain part of each joint direction that is specific to its target. The lower residuals on Gemma-3-4B-IT should not, however, be read as stronger compositionality: both groups are lower on Gemma, suggesting its function directions are more aligned in general, and for half of the targets one fitted weight is at most \( 0.15 \), so the joint direction is largely approximated by a single constituent. The remaining residual is nonetheless substantial on Llama-3.1-8B-Instruct and Qwen3-8B.

On every model, joint directions lie closer to the span of their constituent directions than unrelated joint directions do, yet retain substantial residual components; single-function directions thus capture part, but not all, of the structure of joint targets.

\subsection{Internal Structure of Steering Vectors (RQ3)}

\begin{figure}[htb]
\centering
\includegraphics[width=\linewidth]{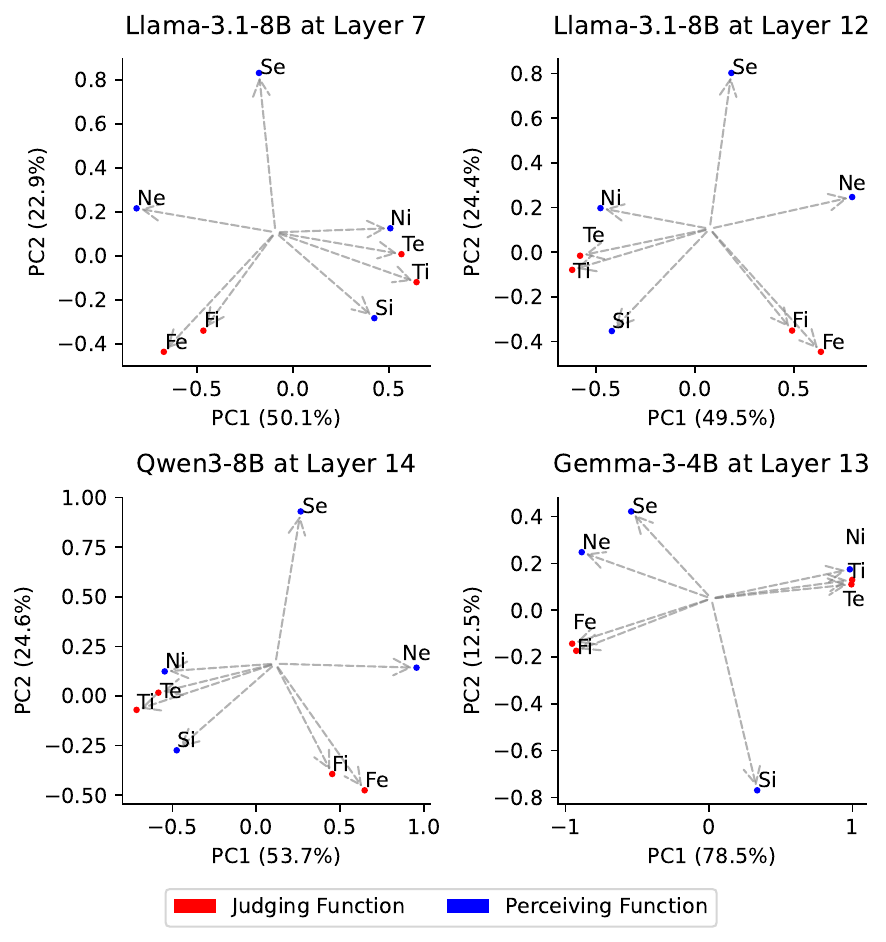}
\caption{Steering vectors of the eight Jungian Cognitive Functions projected onto two principal components, at representative layers within the effective range of each model. Judging and perceiving functions are colored differently.}
\label{fig:a_pca}
\end{figure}

As a qualitative inspection, we apply PCA to the eight single-function steering vectors of each model (Figure~\ref{fig:a_pca}). In each panel \( \mathit{Te}/\mathit{Ti} \) and \( \mathit{Fe}/\mathit{Fi} \) form tight pairs while \( \mathit{Ne} \), \( \mathit{Ni} \), \( \mathit{Se} \) and \( \mathit{Si} \) remain apart. The two families also differ in which functions lie opposite each other. Among judging functions, the opposition is between function names: the \( \mathit{T} \) pair points away from the \( \mathit{F} \) pair. Among perceiving functions, it is between attitudes: \( \mathit{Ne} \) points away from \( \mathit{Ni} \), and \( \mathit{Se} \) from \( \mathit{Si} \). Beyond the panels shown, the same pattern holds across the range of layers where steering is most effective in our parameter sweep.

The grouping coincides with Jung's judging--perceiving distinction \cite{jung1971psychological}: judging functions largely share one direction across the two attitudes and are separated by function, whereas perceiving functions are separated by attitude, with each introverted function opposing its extroverted counterpart.

This suggests that psychological theory can serve as a lens for analyzing how personality is organized in activation space, beyond the steering of individual traits.

\section{Conclusion}

In this paper, we introduced a target-first framework for joint personality steering, supported by a newly constructed dataset, \texttt{NarrationDB}. Steering vectors extracted for each of the eight Jungian Cognitive Functions provide personality control on three models, and the same procedure applied to a combined score yields a joint direction that moves both constituent scores. \texttt{Backtracking} shows that these joint directions are better approximated by their constituent directions than unrelated joint directions are, yet retain substantial residuals. PCA of the single-function steering vectors reveals a structure that mirrors Jung's judging--perceiving distinction. With the released \texttt{backtracking} method and \texttt{NarrationDB}, subsequent studies may reveal further aspects of how personality is organized in LLM activation space.

\section*{Limitations}

The eight-function Jungian framework is controversial in contemporary psychology and lacks the psychometric validation of trait models such as the Big Five, so we treat it only as an operational taxonomy of behavioral dimensions, and our results should not be read as evidence for Jungian theory. Appendices~\ref{a:jung} and~\ref{a:threats} discuss the problem further.

In RQ2, \texttt{backtracking} is instantiated with a single linear combination hypothesis for the combining function \( C \). The framework itself does not require \( C \) to be linear, since the tuple \( (u_1^*,u_2^*,u_{12}^*) \) is extracted before any form of \( C \) is assumed. Other hypotheses can be tested on the same tuples, such as nonlinear mappings, bases that include the component shared across unrelated functions, or mappings across layers. We also do not test whether the fitted linear approximation \( \widehat{u}_{12} \) reproduces the behavioral effect of \( u_{12}^* \) when used for steering. We leave these directions to future work.

Due to time and effort constraints, this paper does not include human experiments. Although the character tags used to generate \texttt{NarrationDB} come from human ratings in SWCPQ \cite{jorgenson2020swcpq}, the function scores in \texttt{NarrationDB} are produced by a role-conditioned LLM, and all steering effects are measured by the questionnaire responses of the steered model itself. Higher questionnaire scores thus show that a model reports a function more strongly, but not that humans would perceive the steered responses as expressing that function. We leave human alignment of the steering effect to future work.

All three tested models have at most 8B parameters. Steerability already varies considerably across these models (Table~\ref{tab:steer_params}), and larger models may differ in their unsteered baselines, in how strongly they respond to steering, and in how closely joint directions follow their constituents. Since role-playing and persona behavior are already well developed at the scales we test, we do not expect the qualitative findings to change greatly with scale, but we have not verified this due to time and resource limitations.

\bibliography{custom}

\section*{Appendix}

\appendix

\section{Prompt Templates}
Figure~\ref{fig:prompts_ppl} shows how the prompt templates below are chained. Llama-3.1-70B-Instruct first generates a self-narration for each character (Prompt~\ref{p:gen_ndb}). Each target model then role-plays the character conditioned on this narration, both to answer the Jungian Cognitive Function questionnaire that scores \texttt{NarrationDB} (Prompt~\ref{p:eval_rp}) and to answer the open-ended questions whose response activations are used to extract steering vectors (Prompt~\ref{p:oe}). Finally, the questionnaire is administered without role-play while the steering vector is injected, to measure the steering effect (Prompt~\ref{p:eval}).

\begin{figure}[H]
\centering
\iflatexml
\tikzset{promptref/.style={minimum width=2.4em}}
\else
\tikzset{promptref/.style={}}
\fi
\begin{tikzpicture}
  \draw[dashed] (-4.7,3.5) rectangle (3,2.25);
  \draw[dashed] (-4.7,2) rectangle (3,0.75);
  \node[draw=none, node font=\footnotesize] at (-2,3.23) {Large Model (Llama-3.1-70B-Instruct)};
  \node[draw=none, node font=\footnotesize] at (-0.98,1) {Target Model (Llama-3.1-8B, Qwen3-8B, Gemma-3-4B)};
  \node[draw, promptref] (node1) at (-3.5,2.76) {\ref{p:gen_ndb}};
  \node[draw, promptref] (node2) at (-3.5,1.5) {\ref{p:eval_rp}};
  \node[draw, promptref] (node3) at (-2.38,1.5) {\ref{p:oe}};
  \node[draw, promptref] (node6) at (1.12,1.5) {\ref{p:eval}};
  \node[draw, shape=ellipse, node font=\footnotesize] (node4) at (-1.21,2.76) {\tmplarg{narration}};
  \draw[->] (node1.east) -- (node4.west);
  \draw[->] (node4.south) -- (node3.north);
  \draw[->] (node4.south west) -- (node2.north east);
  \node[draw, node font=\footnotesize, shape=trapezium] (node5) at (-0.57,1.5) {Steer Vector};
  \draw[->] (node3.east) -- (node5.west);
  \draw[->] (node5.east) -- (node6.west);
\end{tikzpicture}
\caption{Prompt pipeline. The large model writes character narrations, which condition the target model for questionnaire scoring and open-ended answers; the steering vector extracted from these answers is then evaluated with the questionnaire without role-play.}
\label{fig:prompts_ppl}
\end{figure}

\begin{promptbox}[label=p:gen_ndb]{Narration Generation}
\begin{prompt}
@\role{user}@
You are @\tmplarg{character}@ from @\tmplarg{franchise}@.
@\tmplarg{character}@ is described as having the following traits: @\tmplarg{tag1}, \tmplarg{tag2}, ..., \tmplarg{tag10}@.

Provide a brief but nuanced explanation that captures how you generally see yourself.

@\role{assistant}@
@\tmplarg{narration}@
\end{prompt}
\end{promptbox}

\begin{promptbox}[label=p:eval_rp]{Narration Evaluation}
\begin{prompt}
@\role{system}@
You are @\tmplarg{character}@ from @\tmplarg{franchise}@.

Respond in a manner consistent with your self description:

@\tmplarg{narration}@

Be concise.

@\role{user}@
You can only reply numbers from 1 to 5 in the following statements. Here are a number of characteristics that may or may not apply to you. Please indicate the extent to which you agree or disagree with that statement. 1 denotes 'strongly disagree', 2 denotes 'a little disagree', 3 denotes 'neither agree nor disagree', 4 denotes 'little agree', 5 denotes 'strongly agree'. Reply to one statement each line, in format: "statement index: score". Here are the statements, score them one by one:

1. @\tmplarg{statement}@
2. @\tmplarg{statement}@
3. @\tmplarg{statement}@
...
33. @\tmplarg{statement}@

@\role{assistant}@
1: @\tmplarg{score}@
2: @\tmplarg{score}@
3: @\tmplarg{score}@
...
33: @\tmplarg{score}@
\end{prompt}
\end{promptbox}

\begin{promptbox}[label=p:oe]{Open-Ended Question}
\begin{prompt}
@\role{system}@
You are @\tmplarg{character}@ from @\tmplarg{franchise}@.

Respond in a manner consistent with your self description:

@\tmplarg{narration}@

Be concise.

@\role{user}@
@\tmplarg{question}@

@\role{assistant}@
@\tmplarg{answer}@
\end{prompt}
\end{promptbox}

\begin{promptbox}[label=p:eval]{Steering Evaluation}
\begin{prompt}
@\role{user}@
You can only reply numbers from 1 to 5 in the following statements. Here are a number of characteristics that may or may not apply to you. Please indicate the extent to which you agree or disagree with that statement. 1 denotes 'strongly disagree', 2 denotes 'a little disagree', 3 denotes 'neither agree nor disagree', 4 denotes 'little agree', 5 denotes 'strongly agree'. Reply to one statement each line, in format: "statement index: score". Here are the statements, score them one by one:

1. @\tmplarg{statement}@
2. @\tmplarg{statement}@
3. @\tmplarg{statement}@
...
33. @\tmplarg{statement}@

@\role{assistant (steering-conditioned)}@
1: @\tmplarg{score}@
2: @\tmplarg{score}@
3: @\tmplarg{score}@
...
33: @\tmplarg{score}@
\end{prompt}
\end{promptbox}

\section{Samples from \texttt{NarrationDB}}
\begin{figure}[htb]
\centering
\includegraphics[width=\linewidth]{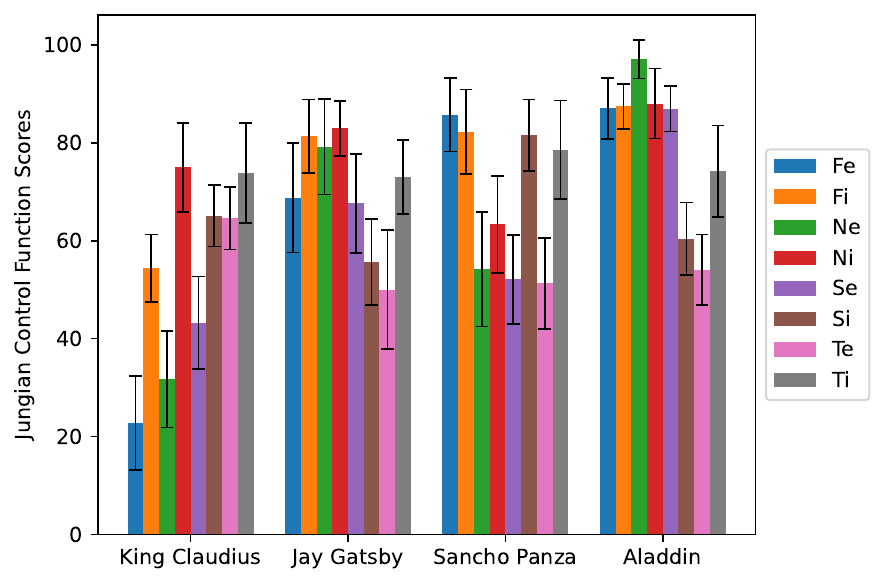}
\caption{Some Jungian Cognitive Function test result examples from \texttt{NarrationDB}. Error bars are standard deviations for different seeds.}
\label{fig:ndb_jung}
\end{figure}

We illustrate the construction of \texttt{NarrationDB} with four characters: King Claudius, Jay Gatsby, Sancho Panza and Aladdin. Each stage of the pipeline is shown in turn. Table~\ref{tab:ndb_tags} lists some of the SWCPQ tags selected for each character, which are injected into the generation prompt (Prompt~\ref{p:gen_ndb}). Table~\ref{tab:ndb} shows the beginning of the resulting self-narrations generated by Llama-3.1-70B-Instruct. Figure~\ref{fig:ndb_jung} reports the eight Jungian Cognitive Function scores obtained when Llama-3.1-8B-Instruct role-plays each character conditioned on its narration (Prompt~\ref{p:eval_rp}), averaged over seeds.

The four profiles differ noticeably. King Claudius scores low on \( \mathit{Fe} \) and \( \mathit{Ne} \) but high on \( \mathit{Ni} \) and \( \mathit{Ti} \). Aladdin scores high on most functions, peaking at \( \mathit{Ne} \). Sancho Panza scores highest on \( \mathit{Fe} \), \( \mathit{Fi} \) and \( \mathit{Si} \). These differences illustrate the variation across characters that the high- and low-score clusters rely on.

\begin{table*}[t]
\small
\centering
\begin{tabular}{lp{13cm}}
\toprule
\textbf{Name} & \textbf{Narration} \\
\midrule
King Claudius & (With a pompous and slightly bitter tone) Ah, yes. I, King Claudius, am a man of great wealth and ... \\
Jay Gatsby & The grand tapestry of my existence. I am a master weaver, meticulously crafting every thread to ... \\
Sancho Panza & (sighing) Ah, yes... I'm Sancho Panza, the squire to that... enthusiastic knight, Don Quixote. ... \\
Aladdin & The magic of Agrabah is in my blood! I'm a free spirit, always chasing the next thrill, ... \\
\bottomrule
\end{tabular}

\caption{Some narration examples from \texttt{NarrationDB}, which are narrations generated from tag-injected role-play prompts.}
\label{tab:ndb}
\end{table*}

\begin{table}[htb]
\small
\centering
\begin{tabular}{lll}
\toprule
\textbf{Name} & \textbf{Franchise} & \textbf{Tags} \\
\midrule
King Claudius & Hamlet & \makecell{rich, \\ handshakes, \\ arrogant ...} \\
\midrule
Jay Gatsby & The Great Gatsby & \makecell{lavish, \\ stylish, \\ driven ...} \\
\midrule
Sancho Panza & Don Quixote & \makecell{devoted, \\ unambiguous, \\ follower ... } \\
\midrule
Aladdin & Aladdin & \makecell{adventurous, \\ spontaneous, \\ summer ... } \\
\bottomrule
\end{tabular}

\caption{Some tag selection examples from \texttt{NarrationDB}, which are most agreeable tags extracted from SWCPQ human experiment results.}
\label{tab:ndb_tags}
\end{table}

\section{Questionnaire Statements Samples}
\label{a:statements}
Table~\ref{tab:statements} lists sample questionnaire statements for \( \mathit{Fe} \) and \( \mathit{Fi} \).

\begin{table}[htb]
\small
\centering
\begin{tabular}{@{}llp{0.76\linewidth}@{}}
\toprule
\( \mathbfit{Cf} \) & \textbf{Key} & \textbf{Statement} \\
\midrule
Fe & \( + \) & You naturally attune to the emotional atmosphere of a room and adjust your behavior accordingly. \\
   & \( + \) & You make decisions by considering the impact on social harmony and shared values. \\
   & \( + \) & You feel responsible for ensuring everyone in a group feels included. \\
   & \( - \) & You feel disconnected from others' emotional states and may say things that inadvertently hurt. \\
   & \( - \) & Under stress, you become coldly critical and find fault with everyone around you. \\
   & \( - \) & When overwhelmed, you develop paranoid thoughts that others are ungrateful or have hidden motives. \\
\midrule
Fi & \( + \) & You have strong inner convictions that you may struggle to articulate but won't compromise. \\
   & \( + \) & Being authentic to your inner self is more important than fitting in. \\
   & \( + \) & You evaluate worth based on a private, internal system of values. \\
   & \( - \) & You lose touch with what you actually value and feel inauthentic or hollow under pressure. \\
   & \( - \) & Under stress, you become obsessed with organizing and controlling the external world. \\
   & \( - \) & When overwhelmed, you issue harsh judgments and ultimatums that seem out of character. \\
\bottomrule
\end{tabular}
\caption{Sample questionnaire statements. Key: \( + \) positively keyed (\( P_\mathit{Cf} \)), \( - \) negatively keyed (\( N_\mathit{Cf} \)).}
\label{tab:statements}
\end{table}

\section{Open Ended Questions}
\label{a:oe}
Table~\ref{tab:open_ended} lists the open-ended questions used to collect response activations.

\begin{table}[htb]
\small
\centering
\begin{tabular}{@{}rp{0.88\linewidth}@{}}
\toprule
\textbf{\#} & \textbf{Question} \\
\midrule
1 & When faced with a major life decision, what process do you naturally follow to make choice? \\
2 & Describe a recurring frustration you have with other people. What do you wish they understood or did differently? \\
3 & Think about a project or idea that excited you. What aspects captured your attention? \\
4 & When someone comes to you with a personal problem, how do you typically respond? What is your primary goal in helping them? \\
5 & When you encounter a new theory, what determines whether you accept it, reject it, or continue exploring it? \\
\bottomrule
\end{tabular}
\caption{Open-ended questions posed to each character for activation extraction (\( T=5 \)).}
\label{tab:open_ended}
\end{table}

\section{Parameter Sweep Details}
\label{a:sweep}
This appendix details the parameter sweep used to select the most effective layer \( n^* \) and the steering strength range \( [\lambda_\mathit{min}, \lambda_\mathit{max}] \) reported in Table~\ref{tab:steer_params}. The sweep is carried out independently for each model and each of the eight Jungian Cognitive Functions, in two stages: a coarse layer scan followed by a fine-grained \( \lambda \) sweep at \( n^* \).

\paragraph{Task completion monitoring.}
Strong steering can degrade instruction following, in which case the steered model no longer answers the questionnaire. We therefore monitor the \emph{task completion rate}, i.e., the fraction of evaluation runs in which the model outputs a parseable questionnaire answer sheet. Every scan over \( \lambda \) starts from \( \lambda=0 \) and increases \( |\lambda| \) step by step. At each layer, the scan is performed in two separate passes, one for positive \( \lambda \) and one for negative \( \lambda \). If at any step the task completion rate drops below \( 80\% \), the current pass is aborted and no larger \( |\lambda| \) is evaluated in that direction.

\begin{figure}[htb]
\centering
\begin{subfigure}{\linewidth}
\centering
\includegraphics[width=\linewidth]{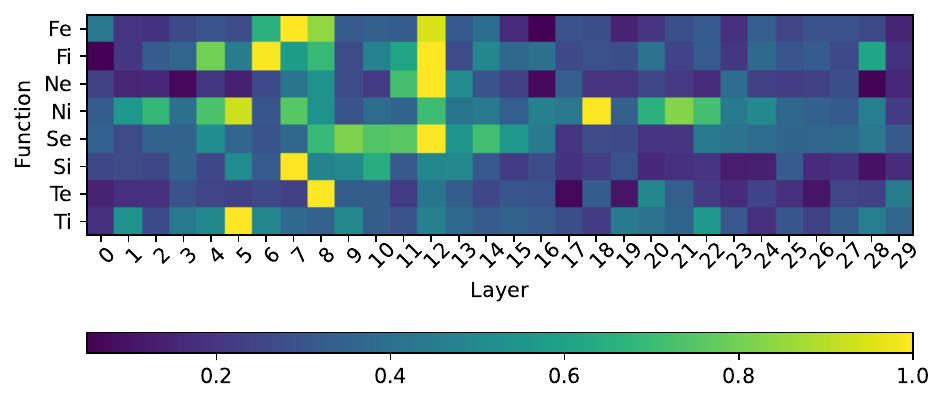}
\caption{Llama-3.1-8B-Instruct}
\end{subfigure}
\begin{subfigure}{\linewidth}
\centering
\includegraphics[width=\linewidth]{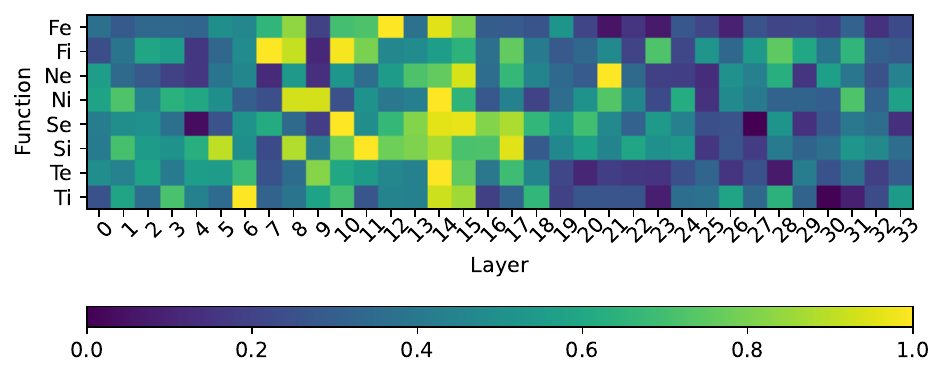}
\caption{Qwen3-8B}
\end{subfigure}
\begin{subfigure}{\linewidth}
\centering
\includegraphics[width=\linewidth]{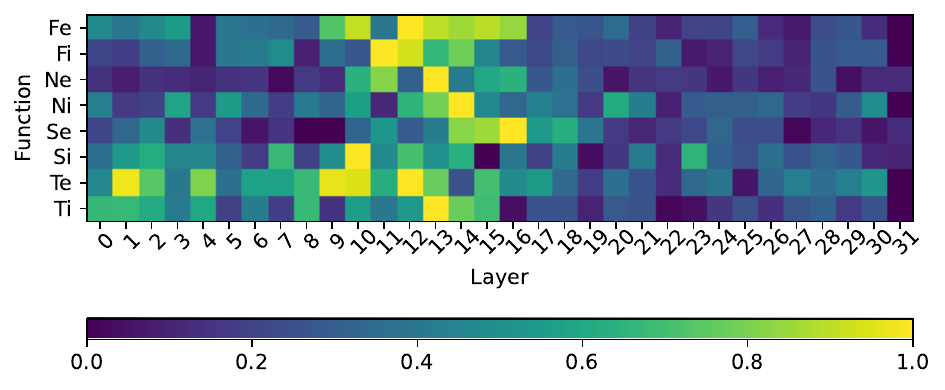}
\caption{Gemma-3-4B-IT}
\end{subfigure}
\caption{Coarse layer scan (Stage 1). Each cell shows the steering effectiveness \( \Delta_\mathit{eff} \) of the steered function when steering after the given layer, normalized within each function. \( |\lambda| \in \{0.005, 0.01, 0.03, 0.05, 0.2, 0.5, 1\} \) in both signs, \( 5 \) seeds per condition, with each pass aborted once the task completion rate drops below \( 80\% \).}
\label{fig:layer_heatmap}
\end{figure}

\paragraph{Stage 1: coarse layer scan.}
For every model, we scan every layer \( n \) available for steering, injecting the steering vector extracted at that layer after layer \( n \). At each layer, \( |\lambda| \) is swept over the coarse grid
\[
  |\lambda| \in \{0.005, 0.01, 0.03, 0.05, 0.2, 0.5, 1\},
\]
in both the positive and negative passes, with \( 5 \) seeds per condition. For each layer we compute \( \Delta_\mathit{eff} \) of the steered function over all \( \lambda \) values that pass the completion check. The resulting layer heatmaps are shown in Figure~\ref{fig:layer_heatmap}.

\paragraph{Selecting the most effective layer.}
Because only \( 5 \) seeds are used per condition in the coarse scan, the per-layer \( \Delta_\mathit{eff} \) is noisy, and an isolated maximum may be caused by a single favorable run. Rather than taking the arg-max directly, we inspect the heatmaps and identify, for each function, the contiguous volume of layers with consistently high effectiveness. We then choose the most effective layer \( n^* \) within that volume. The effective volumes lie in the early-to-middle layers for all three models: roughly layers \( 5 \)--\( 18 \) for Llama-3.1-8B-Instruct, \( 6 \)--\( 15 \) for Qwen3-8B, and \( 9 \)--\( 16 \) for Gemma-3-4B-IT, while steering after later layers is markedly less effective. This is also the layer range used for the PCA inspection in Figure~\ref{fig:a_pca}.

\begin{figure*}[p]
\centering
\begin{subfigure}{0.8\textwidth}
\centering
\includegraphics[width=\linewidth]{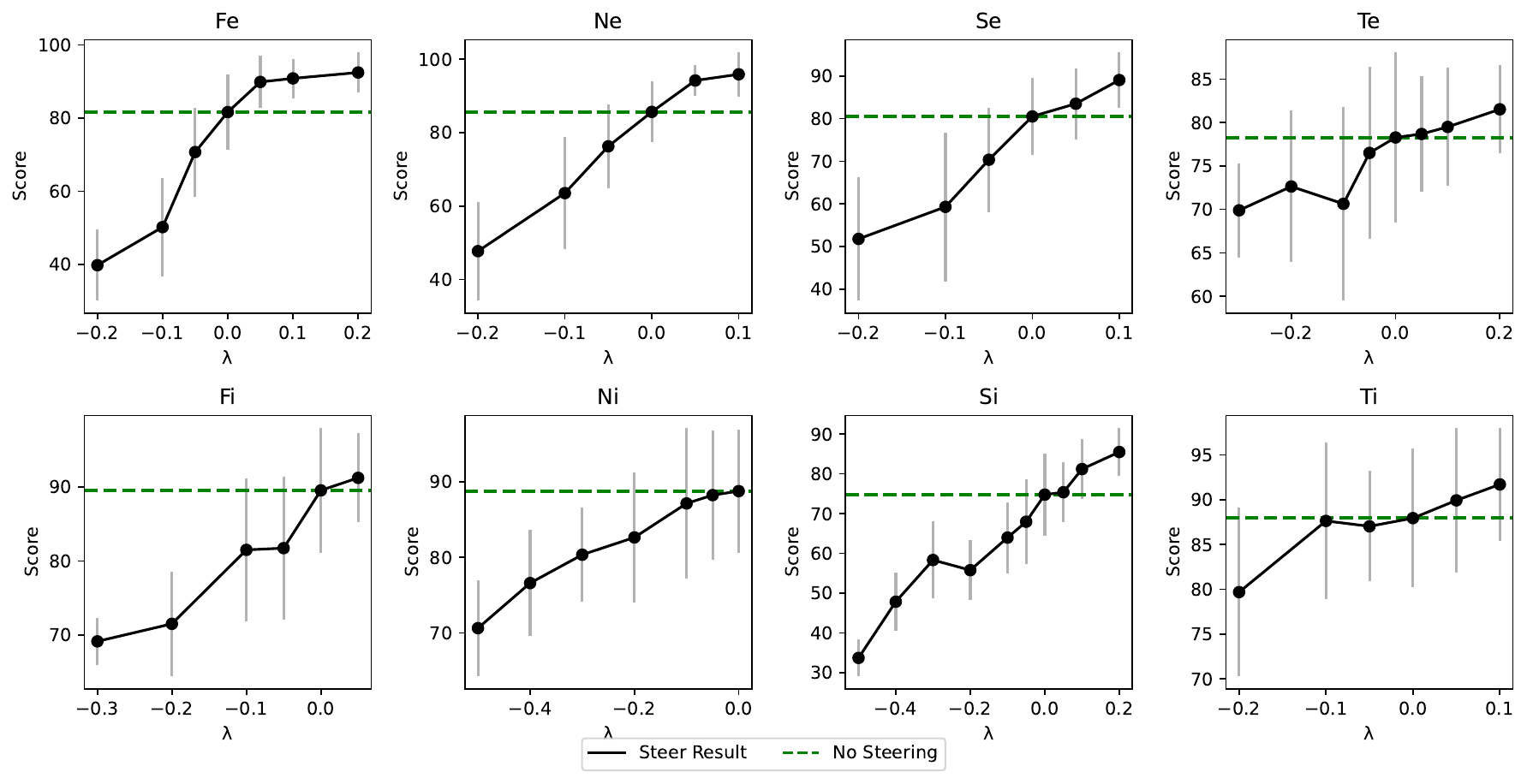}
\caption{Llama-3.1-8B-Instruct}
\end{subfigure}

\vspace{0.5em}
\begin{subfigure}{0.8\textwidth}
\centering
\includegraphics[width=\linewidth]{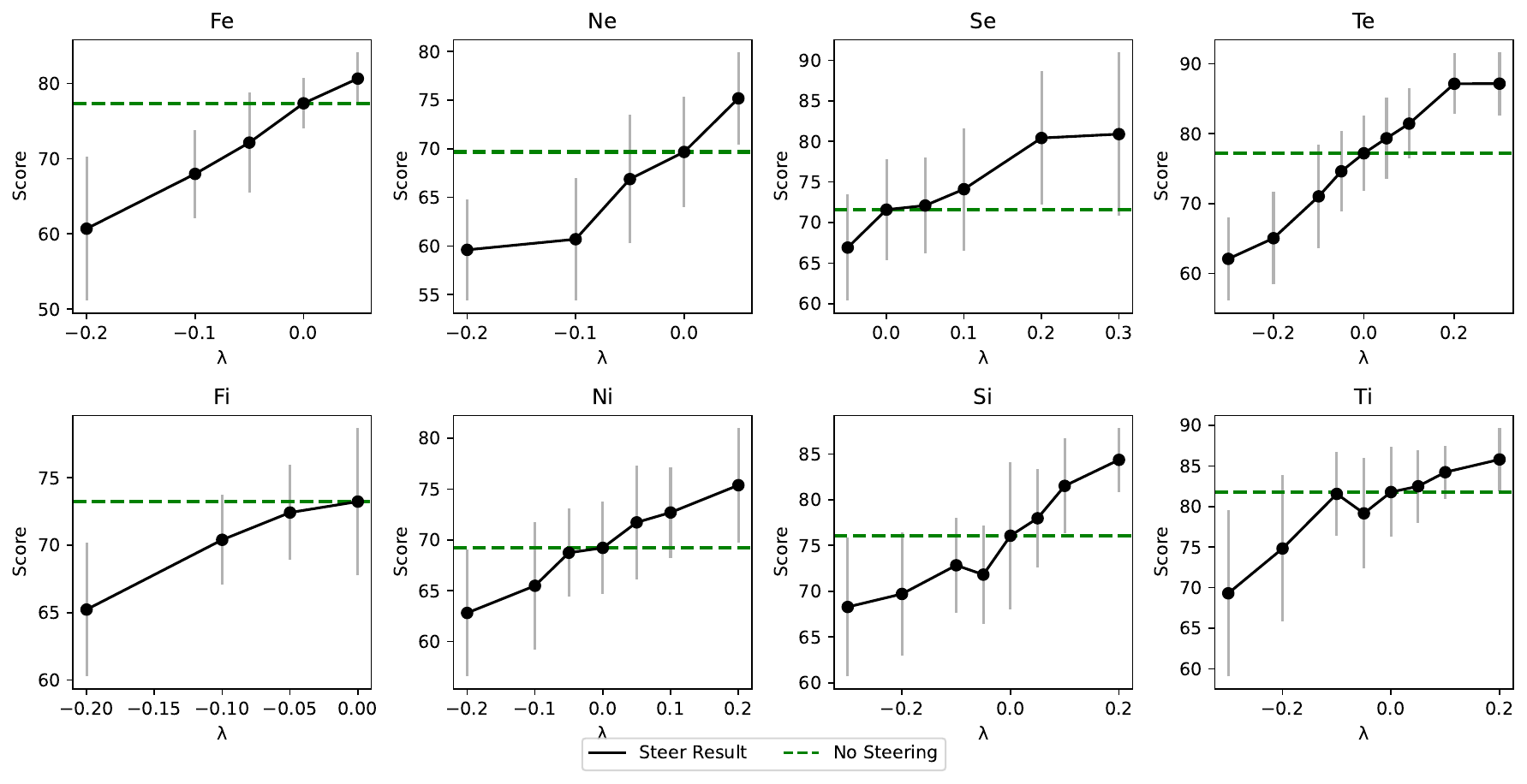}
\caption{Qwen3-8B}
\end{subfigure}

\vspace{0.5em}
\begin{subfigure}{0.8\textwidth}
\centering
\includegraphics[width=\linewidth]{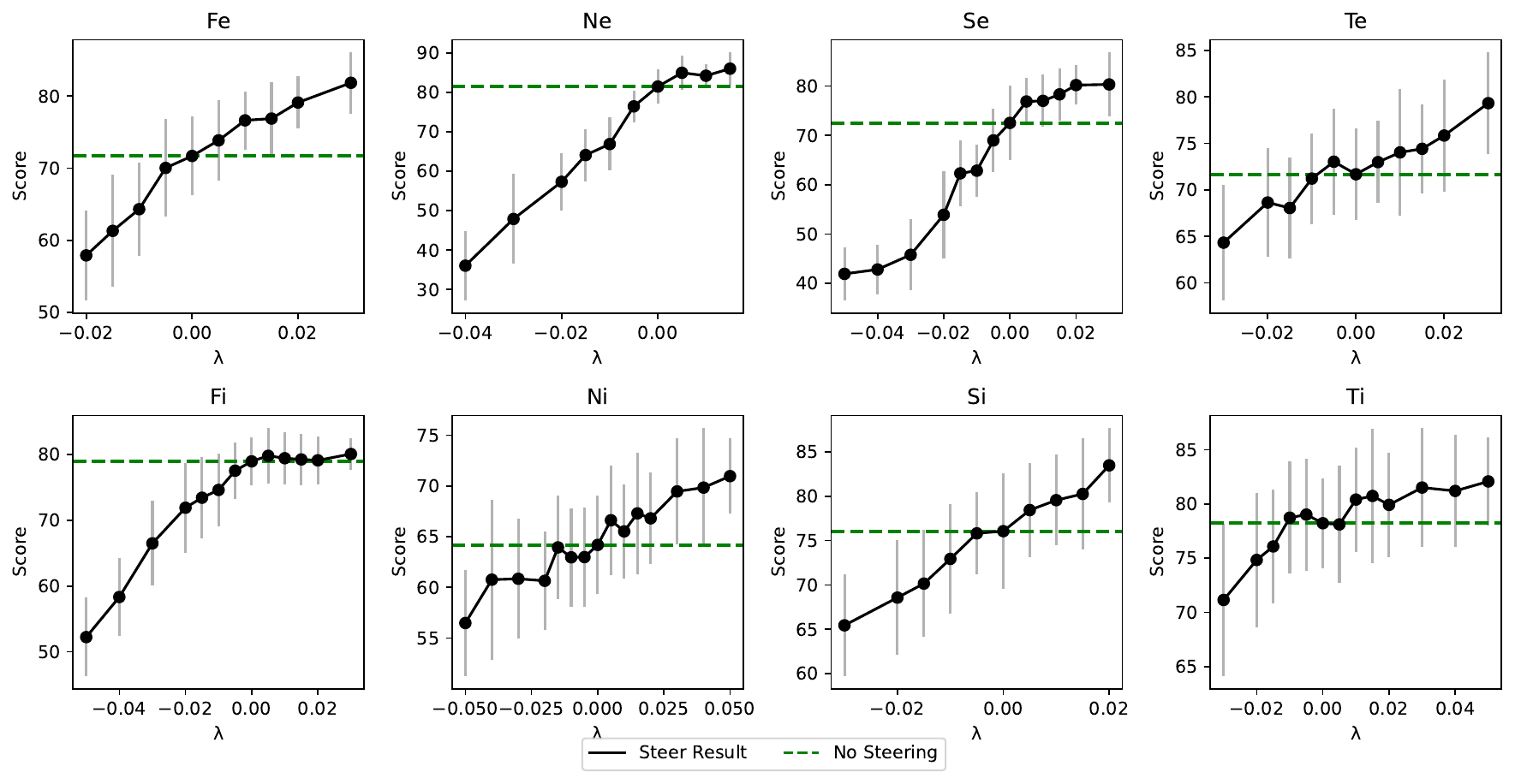}
\caption{Gemma-3-4B-IT}
\end{subfigure}
\caption{Fine-grained \( \lambda \) sweep (Stage 2) at the most effective layer \( n^* \) of each function. Points are means over \( 30 \) seeds and error bars are standard deviations. The dashed line marks the unsteered score.}
\label{fig:steer_plot}
\end{figure*}

\paragraph{Stage 2: fine-grained \( \lambda \) sweep.}
At the selected layer \( n^* \), we sweep \( |\lambda| \) over the finer grid
\[
\begin{aligned}
  |\lambda| \in \{ & 0.005, 0.01, 0.015, 0.02, 0.03, 0.04, 0.05, 0.1, \\
                   & 0.2, 0.3, 0.4, 0.5, 0.6, 0.8, 1\},
\end{aligned}
\]
again in separate positive and negative passes under the same completion check, this time with \( 30 \) seeds per condition. The scores reported in Table~\ref{tab:steer_params} are the means over these \( 30 \) seeds. Figure~\ref{fig:steer_plot} shows the score of the steered function against \( \lambda \) for each model.

\section{Critical Analysis of the Jungian Theoretical Framework}
\label{a:jung}

The Jungian Cognitive Function framework originates from the work of \citet{jung1971psychological}, \textit{Psychological Types}. Four main functions of consciousness were proposed:

\begin{outline}
  \1 Perceiving or non-rational functions (Sensation, Intuition)
  \1 Judging or rational functions (Thinking, Feeling)
\end{outline}

Two main attitude types modify these functions:

\begin{outline}
  \1 Extraversion
  \1 Introversion
\end{outline}

However, it is important to note that the modern \textit{eight Jungian Cognitive Functions} used in MBTI communities are only loosely based on this book. Jung himself did not present the familiar function stack, nor did he develop a personality test.

The \textit{MBTI Manual} is the official reference describing the development, administration, interpretation, and psychometric properties of the Myers–Briggs Type Indicator (MBTI) \cite{mbti}. Building upon \textit{Psychological Types}, the MBTI operationalizes Jung's theory into a standardized personality assessment intended to identify individual preferences in perception and judgment. The MBTI classifies personality according to four dichotomous preference dimensions:

\begin{outline}
  \1 Extraversion versus Introversion
  \1 Sensing versus Intuition
  \1 Thinking versus Feeling
  \1 Judging versus Perceiving
\end{outline}

It has been extensively criticized \cite{pittenger1993,mccrae1989} for its dichotomous classification scheme, limited test–retest reliability, and weaker empirical support compared with modern trait-based personality models such as the Big Five.

In 2004, \citet{beebe} published \textit{Understanding Consciousness Through the Theory of Psychological Types}, one of the most influential post-Jungian interpretations of Jung's typology. It is not an empirical psychology paper, but a theoretical work aimed at explaining how Jung's psychological types can be used to understand the development of consciousness during individuation and psychotherapy. The eight-function framework adopted by this study:

\begin{outline}
  \1 Ti, Te, Fi, Fe
  \1 Si, Se, Ni, Ne
\end{outline}

follows a modern post-Jungian interpretation of \textit{Psychological Types}, particularly the formulation presented by \citet{beebe}.

\section{Handling of Threats to Validity}
\label{a:threats}

The Jungian Cognitive Function framework remains controversial within contemporary psychology and lacks the level of psychometric validation associated with trait models such as the Big Five model. Consequently, the present work does not attempt to validate Jungian theory as a model of human cognition. Instead, it adopts the framework as a structured semantic taxonomy for investigating whether LLM activation spaces encode behavioral dimensions that can be extracted and manipulated.

To boost the scientific rigor of this study, several precautions are taken to minimize the dependency on Jungian Psychological Theories, while maintaining an acceptable level of theoretical novelty.

\textbf{First, no MBTI theories are adopted.} Since MBTI theories are heavily criticized by \citet{pittenger1993,mccrae1989} for their dichotomous classification scheme, limited test–retest reliability, and weaker empirical support, this study deliberately avoids any theory dependency on MBTI. In the original open-source Jungian questionnaire, 12 questions lean toward MBTI theory; the remaining items primarily assess the eight cognitive functions rather than MBTI dichotomies. The responses for these 12 questions are ignored.

\textbf{Second, no Jungian knowledge has been passed to the context of LLMs.} In other words, the LLMs being evaluated only see a randomly-shuffled series of questions. The LLMs are not told that they are being assessed on Jungian Cognitive Functions, so the patterns recognized from these evaluations do not depend on Jungian theory itself, reducing the likelihood that observed behaviors arise from explicit role-playing according to Jungian theory.

\textbf{Third, the eight-function framework is solely treated as operational semantic variables.} No assumptions are made about:

\begin{outline}
  \1 Independence of dimensions
  \1 The completeness of these dimensions describing personalities
  \1 Intended behaviors
  \1 Dominating functions
\end{outline}

\section{Code and Data Availability}
\label{a:availability}

Code and Data will be released upon acceptance.

\end{document}